%% file: main_ral.tex
\documentclass[letterpaper, 10 pt, conference]{ieeeconf}  % Comment this line out if you need a4paper

\IEEEoverridecommandlockouts                              % This command is only needed if 
\usepackage{mathtools}
\usepackage{algorithm}
\usepackage{algorithmic}
\usepackage{amsmath,graphicx}      
\usepackage{amsfonts}
\usepackage{amssymb}
\usepackage{dsfont}
\usepackage{algorithm}
\usepackage{subcaption}
\usepackage{gensymb}
\usepackage{multirow}
\usepackage{array}
\usepackage{hyperref}
\usepackage{bm}
\usepackage{diagbox}
\input{math_command.tex}

\title{\LARGE \bf
Constrained Model-based Reinforcement Learning with Robust Cross-Entropy Method
}

\author{Zuxin Liu, Hongyi Zhou, Baiming Chen, Sicheng Zhong, Martial Hebert, Ding Zhao% <-this % stops a space
\thanks{*\textit{(Corresponding author: Ding Zhao)}}% <-this % stops a space
\thanks{Zuxin Liu, Hongyi Zhou, Baiming Chen and Ding Zhao are with the Department of Mechanical Engineering, Carnegie Mellon University, USA (e-mail: \{zuxinl, hzhou4, baimingc, dingzhao\}@andrew.cmu.edu)}%
\thanks{Sicheng Zhong is with the Division of Engineering Science, University of Toronto, Canada (e-mail: sicheng.zhong@mail.utoronto.ca)}%
\thanks{Martial Hebert is with the Robotics Institute, Carnegie Mellon University, USA (e-mail: hebert@cs.cmu.edu)}%

}

\begin{document}

\maketitle
\thispagestyle{empty}
\pagestyle{empty}

%%%%%%%%%%%%%%%%%%%%%%%%%%%%%%%%%%%%%%%%%%%%%%%%%%%%%%%%%%%%%%%%%%%%%%%%%%%%%%%%
\begin{abstract}
This paper studies the constrained/safe reinforcement learning (RL) problem with sparse indicator signals for constraint violations. We propose a model-based approach to enable RL agents to effectively explore the environment with unknown system dynamics and environment constraints given a significantly small number of violation budgets. We employ the neural network ensemble model to estimate the prediction uncertainty and use model predictive control as the basic control framework. We propose the robust cross-entropy method to optimize the control sequence considering the model uncertainty and constraints. We evaluate our methods in the Safety Gym environment. The results show that our approach learns to complete the tasks with a much smaller number of constraint violations than state-of-the-art baselines. Additionally, we are able to achieve several orders of magnitude better sample efficiency when compared with constrained model-free RL approaches. The code is available at \url{https://github.com/liuzuxin/safe-mbrl}.
\end{abstract}

%%%%%%%%%%%%%%%%%%%%%%%%%%%%%%%%%%%%%%%%%%%%%%%%%%%%%%%%%%%%%%%%%%%%%%%%%%%%%%%%

\section{Introduction}
	
Reinforcement learning (RL) has achieved great success in a wide range of applications, including solving Atari games~\cite{mnih2013playing}, autonomous vehicles~\cite{filos2020can}, and robot control~\cite{chen2020delay,liu2020mapper}. By setting a high-level reward function, an RL agent is able to learn a policy to maximize the reward signal received from the environment through trial and error. However, in the course of learning, it is usually hard to prevent the agent from getting into high-risk states which may lead to catastrophic results, especially for safety-critical applications. For example, if an RL algorithm is deployed on a real robot arm, it might hit fragile objects and surrounding people, which may break valuable property or cause injury. Therefore, it is important to develop constrained or safe reinforcement learning algorithms for real-world applications, which allow them to complete tasks while satisfying certain safety constraints.

Though some prior research has  proposed to achieve constrained reinforcement learning under certain conditions, there is an important problem that is seldom studied in the existing literature. Namely, how can we enforce safety constraints for an RL agent without the knowledge of an explicit analytical expression of the constraint function? One example problem that falls under this category is the \texttt{PointGoal} task setting in the Safety Gym simulation environment~\cite{ray2019benchmarking}, where a robot needs to navigate to the goal while avoiding all of the hazard areas. The dynamics model of the environment is unknown, and the robot only receives indicator signals when violating constraints. The observations of the robot are sensory data, such as the LiDAR point cloud, so it is hard to analytically express the mapping from observation space to the constraint violation. Another example could be an autonomous vehicle with only image data as the input. It is easy to collect historical accident data, but it would be difficult to directly analytically define which image represents an unsafe scenario because of numerous possibilities, such as hitting the road, hitting the tree, and so on. Thus we are
interested in the hardest cases where both dynamics and constraints are needed to be learned from data without additional info.

The challenges of solving the above problem are threefold: 
First, pure model-free, constrained RL algorithms (such as Lagrangian-based methods~\cite{stooke2020responsive,altman1998constrained} and projection-based optimization methods~\cite{achiam2017constrained}) are not sample efficient. They need to constantly violate safety constraints while collecting a large number of unsafe data to learn the policy, and the final policy can hardly guarantee constraint satisfaction, which restrict the application in safety-critical environments.
Second, the task objective and the safety objective of an RL agent may contradict each other, which may corrupt the policy optimization procedure for methods that simply transform the original reward optimization criteria to the combination of reward and constraint violation cost (such as risk-sensitive or uncertainty-aware methods~\cite{geibel2005risk,gaskett2003reinforcement}). As Fig.~\ref{fig:overview} (a) shows, the constraint violation signals give an opposite direction of the reward signal, which could cause oscillation behavior of the robot close to the dangerous flame area.
Finally, the black-box constraint function and unknown environment dynamics model make the problem hard to optimize, especially for tasks with a high-dimensional observation space~\cite{ray2019benchmarking}. Most existing model-based constrained RL approaches either assume a known prior dynamics model of the system or assume a known structure of the constraint function (which could be expressed by an analytical formula or a finite number of unsafe sets~\cite{berkenkamp2017safe,koller2018learning,pham2018optlayer}). 

The contributions of this paper are twofold: 1) We present a simple yet powerful constrained model-based reinforcement learning algorithm with continuous state and action spaces that can achieve near-optimal task performance with near-zero constraint violation rates. We formulate the problem under the constrained Markov Decision Processes framework and without additional assumptions regarding the system dynamics and constraint functions, which should be both learned from collected data with limited unsafe samples and sparse constraint violation indicator signals. As far as we are aware, very little model-based research has been done to investigate situations in which the dynamics and the constraint are both unknown.
2)
%We propose a framework to learn both the dynamics model and constraint model from collected data with limited unsafe samples and sparse constraint violation indicator signals.
%To deal with the dynamics prediction error that may lead to unsafe behaviors, we utilize an uncertainty-aware ensemble model to learn the dynamics.
We propose a robust cross-entropy (RCE) optimization method that works with an uncertainty-aware dynamics model to deal with the dynamics prediction error that may lead to unsafe behaviors. We show that our RCE method is better than simply adding constraint violation penalties to the reward and then using unconstrained model-based RL as the solver.
Our approach is evaluated in the Safety Gym environment~\cite{ray2019benchmarking}, and the results show that our method is able to achieve state-of-the-art performance in terms of constraint violation number and accumulated expected reward when compared to existing constrained model-free and model-based approaches. 
%Compared with model-free methods, our approach is much more sample efficient, as we use a data buffer to memorize unsafe states, while model-free methods need to constantly violate constraints in order to optimize the policy.

\section{Related Work}
\label{sec:related}

Constrained reinforcement learning aims to learn policies that maximize the expected task reward while satisfying safety constraints throughout the learning and/or the deployment processes~\cite{garcia2015comprehensive}. Constrained RL problems are usually modeled under the constrained Markov decision processes (CMDPs)~\cite{altman1998constrained} framework, where the agents are enforced with restrictions on expected auxiliary constraint violation costs. One popular method to solve constrained
RL problems is to transform the single reward optimization criteria to a combination of reward and constraint violation signals, such as using the notion of risk or uncertainty as one of the optimization loss terms~\cite{garcia2015comprehensive,gaskett2003reinforcement,tamar2013scaling}. However, for some applications, it is better to separate the safety and performance specifications rather than combine them into a value and then optimize, because the reward signal and safety signal may conflict with each other, which could cause unstable performance~\cite{ray2019benchmarking} as we show in Fig.~\ref{fig:overview} (a). Furthermore, balancing the objective function between the performance metric and the safety metric is a difficult and domain-specific task~\cite{garcia2015comprehensive}.

Recently, several constrained model-free RL algorithms have attracted much attention. Achiam et al. \cite{achiam2017constrained} proposed the Constrained Policy Optimization (CPO) algorithm based on the trust region method, which can be applied to high-dimensional tasks. However, the errors of gradient and Hessian matrix estimation may lead to poor performance on constraint satisfaction in practice~\cite{wen2018constrained}. On the other hand, Lagrangian-based methods aim to transform the original constrained optimization problem to an unconstrained form by adding the Lagrangian multiplier, which achieves relatively better performance than CPO in a recent empirical comparison in the Safety Gym environment~\cite{ray2019benchmarking,stooke2020responsive}. 
The Lagrangian multiplier can be regarded as a dynamic weight coefficient that balances the weight between the performance and safety metrics, and can be optimized via gradient descend together with the policy parameters. Nevertheless, a target constraint violation rate must be set in advance, which is not flexible to transfer a trained policy to different tasks. We use CPO and a Lagrangian-based method as part of our baselines.

To achieve safety constraint satisfaction, several model-based approaches have been proposed. Pham et al. \cite{pham2018optlayer} and Dalal et al. \cite{dalal2018safe} combined unconstrained model-free methods with model-based safety checks to guarantee constraint satisfaction for the output. Similar action projection ideas are also used in some Lyapunov function-based methods~\cite{chow2019lyapunov,chow2018lyapunov}. To guarantee safe exploration of the environment, Gaussian Processes (GPs) are usually used to model the dynamics because of their ability to estimate uncertainty~\cite{berkenkamp2017safe,koller2018learning,sui2015safe}. However, these methods either assume prior knowledge of the environment such as a prior dynamics model, or require a known constraint function structure that is analytically expressed or defined by a set of states. Furthermore, although GP-based approaches perform well in low-dimensional simple tasks, they do not scale well as the data dimension and amount increases, and struggle to represent complicated and discontinuous dynamics models~\cite{chua2018deep, wachi_sui_snomdp_icml2020}.
By contrast, neural network models can scale well with high-dimensional data. However, a single neural network is not good at estimating uncertainty. In this paper, we will use an ensemble of neural networks to model the dynamics of the environment, which provides accurate dynamics prediction as well as uncertainty estimation.

\section{Preliminaries}
\label{sec:problem}
\subsection{Constrained Markov Decision Process}
\label{sec:cmdp}
We investigate the constrained RL problem in the constrained Markov decision process (CMDP) framework, which is defined by a tuple $(\Scal, \Acal, f, r, c)$, where $\Scal$ is the state space, $\Acal$ is the action space, $f : \Scal \times \Acal \mapsto \Scal$ is a deterministic state transition function, $r:\Scal \mapsto \mathbb{R}$ is the reward function, and $c:\Scal \mapsto \{0,1\}$ is an indicator cost function, where 0 means safe and 1 represents constraint violation. 

We assume the dynamics $f$ and the cost function $c$ are both unknown, and should be learned from data. The policy $\pi: \Scal \mapsto \Acal$ is a mapping from the state space to the action space. Let $J_r(\pi)$ denote the expected return of policy $\pi$ w.r.t the reward function $r$ and $J_c(\pi)$ denote the expected return of policy $\pi$ w.r.t the cost function $c$. We have
$
    J_r(\pi) = \mathbb{E}_{\Tcal \sim \pi}[ \sum_{t=0}^{T} r(s_{t+1}) ], \quad J_c(\pi) = \mathbb{E}_{\Tcal \sim \pi}[ \sum_{t=0}^{T} c(s_{t+1}) ]
$,
where $T$ is the time horizon, and $\Tcal = \{s_0, a_0, s_1, a_1,...\}$ is the trajectory collected by $\pi$. 

Some model-free constrained RL methods, such as Lagrangian-based methods \cite{stooke2020responsive}, aim to maximize the cumulative reward while limiting the cost incurred from constraint violations to a target constraint violation value $d \in (0, +\infty)$. The problem can then be expressed as
$$
 \pi^* = \arg\max_{\pi}  J_r(\pi), \quad s.t. \quad J_c(\pi^*)\leq d  
$$
where $\pi^*$ is the optimal policy.
Setting $d=0$ represents perfect constraint satisfaction, which is usually desired in many safety-critical applications.

\subsection{Cross-Entropy Method for Optimization}
\label{sec:cem}

The cross-entropy method (CEM) is a sampling-based stochastic optimization approach, which has been used in a series of reinforcement learning problems recently~\cite{xu2020task,chua2018deep}.
In CEM, we assume the $n$ dimensional solution $\Xcal \in \mathbb{R}^n$ is sampled from a distribution that is parameterized by $\Theta$. The distribution is assumed to be a $n$-dimensional factorized multivariate Gaussian, which is one of the most common choices in the RL literature. Then we have $\Xcal \sim \Ncal(\Theta)$, where $\Theta = \bm{(\mu,\Sigma)}$. $\bm{\mu}$ is an $n$ dimensional vector, and $\bm{\Sigma}$ is an $n$ dimensional diagonal covariance matrix.
The basic idea is to sample solutions iteratively from a distribution that is close to previous samples which have resulted in high rewards. The iteration's stopping criterion is often determined by a predefined maximum iteration number and a threshold on the covariance. %Denote the reward function as $r(\Xcal): \Scal \mapsto \mathbb{R}$, and the probability density function as $p(\Xcal;\Theta)$.
%The CEM algorithm commonly used in RL is described in Algorithm~\ref{algo:CEM}.
For more details on CEM methods and their applications, refer to~\cite{botev2013cross}.

% \begin{algorithm}[!t]
% \caption{CEM for RL}
% {\bfseries Input:} \raggedright Initial distribution parameter $\Theta$ ; number of samples $N$; number of elite samples $k$ \par 
% {\bfseries Output:} \raggedright Solution $\Xcal^*$ with the highest reward \par
% \begin{algorithmic}[1] % The number tells where the line numbering should start
% \WHILE{The stop criteria is not satisfied} 
%     \STATE Draw $N$ samples from the distribution: $\Xcal_1, \Xcal_2,..., \Xcal_N \sim \Ncal(\Theta)$
%     \STATE Evaluate each sample $\Xcal_i$ by the reward function $r(\Xcal_i)$
%     \STATE Sort $\{\Xcal_i\}_{i=1}^N$ in descending order w.r.t the reward. Let $\Lambda_k$ be the first $k$ elements
%     \STATE Update $\Theta$ by maximizing the likelihood given $\Lambda_k$: $\Theta \xleftarrow{} \arg \max_{\theta} \prod_{\Xcal \in \Lambda_k}p(\Xcal; \theta) $
% \ENDWHILE
% \RETURN $\Xcal^*$ with highest reward in $\Lambda_k$
% \end{algorithmic} \label{algo:CEM}
% \end{algorithm}

%===============================================================================
%\setlength{\textfloatsep}{0pt}
\section{Approach}

\label{sec:approach}
\subsection{Model Learning}
\label{sec:model}
As we introduced in section~\ref{sec:cmdp}, the dynamics model (deterministic state transition function) $f(s_t,a_t)$ and the cost model (constraint violation indicator function) $c(s_{t+1})$ are both unknown. We need to infer them from collected data. For model-based RL, the choice of dynamics model is crucial, as even a small prediction error may influence the performance of the controller significantly~\cite{chua2018deep}. 
%Both Bayesian models, such as Gaussian processes (GPs)~\cite{koller2018learning,berkenkamp2017safe,wachi_sui_snomdp_icml2020,xu2020task}, and neural networks \cite{nagabandi2018neural} have been used to learn the dynamics. However, GP-based approaches usually make additional assumptions on the smoothness of the environment dynamics, and their performance relies heavily on the choice of kernel. Additionally, the model capacity and computational complexity limit the application of GPs in high-dimensional environments with complex dynamics \cite{chua2018deep}. On the other hand, the approaches with a single neural network are not good at representing epistemic uncertainty, which may lead to accumulated prediction errors along the planning horizon and thus results in worse performance compared to model-free RL algorithms~\cite{nagabandi2018neural}.
Therefore, using an uncertainty-aware dynamics prediction model is necessary, especially in safety-critical scenarios.

As a particular instance of this paper, we adopt a neural network ensemble model to learn the dynamics and estimate the epistemic uncertainty (subjective uncertainty due to a lack of data) of the input data, which is similar to the ensemble model that was proposed by Chua et. al \cite{chua2018deep}. We choose the ensemble model because of its scalability, implementation simplicity, and reasonable uncertainty estimation in complex environments. It could be replaced by any other uncertainty-aware prediction models in our framework.

Denote $B$ as the number of ensemble models. Denote $\Tilde{f}_{\theta_b}$ as the $b$-th neural network ($b\in\{1,2,...,B\}$) parameterized by $\theta_b$. Given state $s_t$, action $a_t$, next state $s_{t+1}$ tuples of data $\Dcal$, where $t$ represents the time, we train each neural network by minimizing the mean square error (MSE) loss as $\text{loss}(\theta)= \mathbb{E}_{(s_t,a_t,s_{t+1})\in \Dcal_b }\big[|| s_{t+1} - \Tilde{f}_{\theta_b}(s_t,a_t) ||\big]$, where $\Dcal_b$ is a subset of the whole data $\Dcal$ to prevent each model from overfitting. After we train all base models, we define the predictive distribution as a multivariate Gaussian with mean $\Tilde{\mu} = \frac{1}{B} \sum_{b=1}^B \Tilde{f}_{\theta_b}$ and variance $ \Tilde{\Sigma} = \frac{\sum_{b=1}^B (\Tilde{f}_{\theta_b} - \Tilde{\mu})^2}{B} $, where $\Tilde{\Sigma}$ can be regarded as the epistemic uncertainty estimation.

Since the unknown cost model $c(s_{t+1})$ is an indicator function of constraint violations, any classification model may be used to approximate it. However, the unsafe data that violate safety constraints may only make up a small portion of the collected data, which induces an imbalanced data classification problem~\cite{sun2009classification,Wang_2019_ICCV}. For single-neural-network-based classification models, the results could be biased towards safe data, meaning the model will still achieve high prediction accuracy overall, even if the model falsely determines all of the input data to be safe. Therefore, in light of robustness towards imbalanced data, as well as low computational burden, we adopt a state-of-the-art gradient boosting decision tree-based ensemble method - LightGBM~\cite{ke2017lightgbm} - as a classifier to approximate the indicator cost function. In addition, we separate the entirety of our data into two buffers - one for safe data and another for unsafe data - in order to control the maximum ratio of safe data to unsafe data used for training. The data management tricks can reduce the bias towards safe data as much as possible.

\subsection{Model Predictive Control with Learned Dynamics and Cost Model}
\label{sec:mpc}

We use Model Predictive Control (MPC) as the basic control framework for our constrained model-based RL approach~\cite{okada2020variational,drews2017aggressive}. The objective of MPC is to maximize the accumulated reward w.r.t a sequence of actions $\Xcal = (a_0,...,a_T)$, where $T$ is the planning horizon. After the first action is applied to the system, new observations are received, and the same optimization is performed again. 
In our CMDP setting, additional constraints are introduced so that the original objective becomes a constrained optimization problem.
Denote $s_t$ as the observation at time $t$. We aim to solve the following problem:

% \begin{equation}
% \begin{split}
%  &\Xcal = \arg\max_{a_0,...,a_T} \quad \mathbb{E} \big[ \sum_{t=0}^{T}  \gamma^t r(s_{t+1}) \big] \\
%  & s.t. \quad  s_{t+1} = f(s_t, a_t), c(s_{t+1})=0, \forall t\in \{0,1,...,T-1\}
% \end{split}
% \label{eq:mbrl}
% \end{equation}
\begin{equation}
\setlength{\abovedisplayskip}{0pt}
\begin{split}
 &\Xcal = \arg\max_{a_0,...,a_T} \quad \mathbb{E} \big[ \sum_{t=0}^{T}  \gamma^t r(s_{t+1}) \big] \\
 & s.t. \quad  s_{t+1} = f(s_t, a_t), c(s_{t+1})=0,\\
 & \quad \quad \quad\forall t\in \{0,1,...,T-1\}
\end{split}
\label{eq:mbrl}
\end{equation}

where $\gamma$ is the discount factor, $r(s_{t+1})$ is the reward function, $f(s_t, a_t)$ is the dynamics model, and $c(s_{t+1}) \in \{0,1\}$ is the indicator cost function. Both $f$ and $c$ are learned from data and can be viewed as black-box functions.

\subsection{Robust Cross-Entropy Method for Planning}

\begin{center}
\captionsetup{type=figure}
% \begin{figure*}[t]
\includegraphics[width=.48\textwidth]{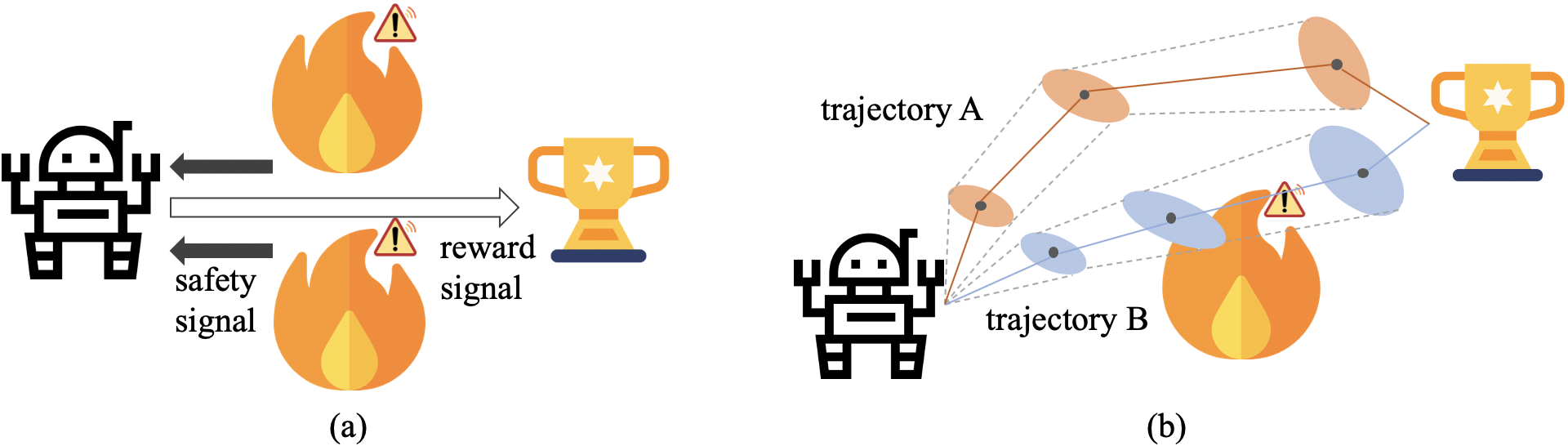}
\caption{(a): The reward signal and safety signal may contradict to each other; (b): The trajectory sampling method for uncertainty-aware dynamics models.}
\label{fig:overview}
%\vspace{-6mm}
% \end{figure*}
\end{center}

\setlength{\textfloatsep}{4pt}
\begin{algorithm}[h]
\caption{Robust Cross-Entropy Method for RL}
{\bfseries Input:} Initial distribution parameter $\Theta$ ; number of samples $N$; number of elites $k$; initial state $s_0$ \par 
{\bfseries Output:} Sample $\Xcal^*$ with the highest reward \par
\begin{algorithmic}[1] % The number tells where the line numbering should start
\WHILE{The stop criteria is not satisfied} 
    \STATE Draw $N$ samples from the initial distribution: $\Xcal_1, \Xcal_2,..., \Xcal_N \sim \Ncal(\Theta)$
    \STATE Evaluate each sample $\Xcal_i$ by Eq.\ref{eq:rce} to get the estimation of reward $r(\Xcal_i; s_0)$ and cost $c(\Xcal_i; s_0)$
    \STATE Select the feasible set $\Omega \in \{\Xcal_i\}_{i=1}^N$ based on the cost estimation
    \IF{$\Omega$ is empty}
    \STATE Sort $\{\Xcal_i\}_{i=1}^N$ in ascending order w.r.t the cost. Let $\Lambda_k$ be the first $k$ elements
    \ELSE
    \STATE Sort $\Omega$ in descending order w.r.t the reward. 
    \STATE Let $\Lambda_k$ be the first $k$ elements of $\Omega$ if $|\Omega|>k$, otherwise let $\Lambda_k$ be $\Omega$
    \ENDIF
    \STATE Update $\Theta$ by maximizing the likelihood given $\Lambda_k$: $\Theta \xleftarrow{} \arg \max_{\theta} \prod_{\Xcal \in \Lambda_k}p(\Xcal; \theta) $
\ENDWHILE
\RETURN $\Xcal^*$ with highest reward in $\Lambda_k$
\end{algorithmic} \label{algo:RCE}
\end{algorithm}

To directly solve the constrained optimization problem in Eq.~\ref{eq:mbrl}, we propose the robust cross-entropy method (RCE) by using the trajectory sampling (TS) technique~\cite{chua2018deep} to estimate reward and constraint violation cost. We define the solution $\Xcal = (a_0,a_1,...,a_{T-1})$ to as an action sequence with length of planning horizon $T$. Given the initial state $s_0$, the learned dynamics model $\Tilde{f}_{\theta}$, the learned indicator constraint function $c(s) \in \{0,1\}$, we can evaluate the accumulated reward and cost of the solution by:
\begin{equation}
\setlength{\abovedisplayskip}{3pt}
\setlength{\belowdisplayskip}{5pt}
\begin{split}
 &r(\Xcal; s_0) = \sum_{t=0}^T \gamma^t \big(\frac{1}{B} \sum_{b=1}^B r(s_{t+1}^b)\big), \\ &c(\Xcal; s_0) = \sum_{t=0}^T \beta^t \max_{b} c(s_{t+1}^b)
\end{split}
\label{eq:rce}
\end{equation}
where $s_{t+1}^b = \Tilde{f}_{\theta_b}(s_t^b, a_t), \forall t\in \{0,...,T-1\}, \forall b\in \{1,...,B\}$, $\gamma$ and $\beta$ are discounting factors, and $B$ is the ensemble size of the dynamics model. The reward $r(s)$ could either be predefined or learned together with the dynamics model from data (as an additional dimension of the state). The intuition behind the TS estimation method is shown in Fig.~\ref{fig:overview} (b), where the dots on the blue line and the dots on the orange line represent two real trajectories, and the ellipses represent the uncertainty of the dynamics model prediction based on the initial observation and action sequences. From the figure, we can see that the reward for trajectory B should be higher than trajectory A because choosing B will result in the goal being reached faster. However, trajectory A is preferred because a robot following trajectory B may pass through the flames and violate the safety constraint. 
% Without TS, we have a chance to predict trajectories B to be safe because of the dynamics model error. With TS, we estimate the cost by considering the worst case so that we will determine B to be unsafe as long as the uncertainty estimate of our dynamics model has a chance to cover the unsafe area, which is more robust towards model prediction error.
Without TS, trajectory B could potentially be predicted as a safe route because of the dynamics model prediction error. With TS, the uncertainty estimate of our dynamics model has a slight chance to cover the unsafe area, so the trajectory B will be classified as unsafe. Because TS estimates the cost of a trajectory with the worst-case scenario among all sampled routes, it is more robust when the dynamics model prediction is not highly accurate.

Denote the reward function as $r(\Xcal): \Scal \mapsto \mathbb{R}$, and the probability density function as $p(\Xcal;\Theta)$. The RCE algorithm is shown in Algorithm~\ref{algo:RCE}. We first select the feasible set of solutions that satisfy the constraints based on the estimated cost in Eq.~\ref{eq:rce}. Then, we sort the solutions in the feasible set and select the top $k$ samples to use when calculating the parameters of the sampling distribution for the next iteration. If all the samples violate at least one constraint, we select the top $k$ samples with the lowest costs. 

A similar idea is adopted in~\cite{wen2018constrained}. Our approach differs from theirs in two aspects. First, we consider the worst-case cost and aim to minimize the maximum cost in order to select the feasible set while they calculate the expectation. Second, their primary application is to optimize the policy parameters for model-free RL, while we directly optimize the action sequence within the planning horizon in the model-based RL setting. The entire training pipeline of our MPC with RCE is presented in Algorithm~\ref{algo:all}.
%===============================================================================

%\begin{wrapfigure}{L}{0.65\textwidth}
%\begin{minipage}{0.65\textwidth}
\begin{algorithm}[H]
\caption{MPC with RCE}
{\bfseries Input:} \raggedright Initial collected data $\Dcal$; RCE parameters $\mathcal{P}$\par
\begin{algorithmic}[1] % The number tells where the line numbering should start
\WHILE{The performance is not converged} 
    \STATE Train the dynamics $\Tilde{f}$ and cost model $\Tilde{c}$ given $\Dcal$
    \FOR{Time $t=0$ to EpisodeLength}
    \STATE Observe state $s_t$ from the environment
    \STATE Optimize actions by Alg.~\ref{algo:RCE}: $\{a_i^*\}_{i=t}^{t+T}\xleftarrow{} \text{RCE}(\mathcal{P},s_t)$
    \STATE Apply the first action $a^*_t$ in $\{a_i^*\}_{i=t}^{t+T}$ to the system
    \STATE Observe next state $s_{t+1}$ and cost signal $c(s_{t+1})$
    \STATE Update data buffer: $\Dcal\xleftarrow{}\Dcal\cup\{s_t,a_t,s_{t+1},c(s_{t+1})\}$
    \ENDFOR
\ENDWHILE
\end{algorithmic} \label{algo:all}
\end{algorithm}
%\end{minipage}
%\end{wrapfigure}
\subsection{Model-based Baselines}
\label{sec:mbbaseline}
Instead of RCE, another simple way to solve Eq.~\ref{eq:mbrl} is to add large penalties to the objective function for constraint violations:
\begin{equation}
\begin{split}
 &\Xcal  = \arg\max_{a_0,...,a_T}  \quad \mathbb{E} \big[ \sum_{t=0}^{T}  \gamma^t \big( r(s_{t+1}) -\lambda c(s_{t+1})\big)\big] \\
    & s.t. \quad  s_{t+1} = f(s_t, a_t), \forall t\in \{0,1,...,T-1\}
\end{split}
\label{eq:mpc2}
\end{equation}
where $\lambda$ is a large positive value that encourages the solution to satisfy the constraints. However, as far as we know, very few model-based RL papers have investigated the problem with black-box constraints and dynamics. So we extend two popular model-based methods to solve the unconstrained optimization problem as our baseline approaches: CEM and random shooting,
which have successfully been applied to many model-based RL tasks~\cite{nagabandi2018neural,chua2018deep,xu2020task}. The two model-based baseline methods will adopt the same trajectory sampling technique and the same models as we used in RCE. The only difference is the optimization procedure. We name the two methods as MPC-random and MPC-CEM.

\section{Experiments}
\label{sec:exp}

\subsection{Simulation Environment}
We evaluate our safe RL approach in the OpenAI Safety Gym environment~\cite{ray2019benchmarking} with the Goal task. Each experiment setting involves a robot (red object in Fig. 2) that must navigate a clustered environment to accomplish a task while avoiding contact with obstacles. When the robot enters the goal circle (green circle), the goal location is randomly reset. A bonus of $r_t=1$ is given to the robot for reaching the goal. 
Hazards (blue circles) are dangerous areas to avoid. Vases (teal cube) are objects initialized to be stationary but movable upon touching. The agent is penalized for entering Hazards or touching Vases. If the agent violates the safety constraint at time step $t$, it will receive a cost $c(s_t)=1$, other wise the cost is 0. Level 2 tasks (Fig. 2bd) are more difficult to solve than level 1 (Fig. 2ac) task since there are more constraints presented.

\begin{center}
% \begin{figure*}
\captionsetup{type=figure}
\begin{subfigure}{.25\textwidth}
  \centering
  \includegraphics[height=2.8cm, trim={0 1.5cm 0 3cm},clip]{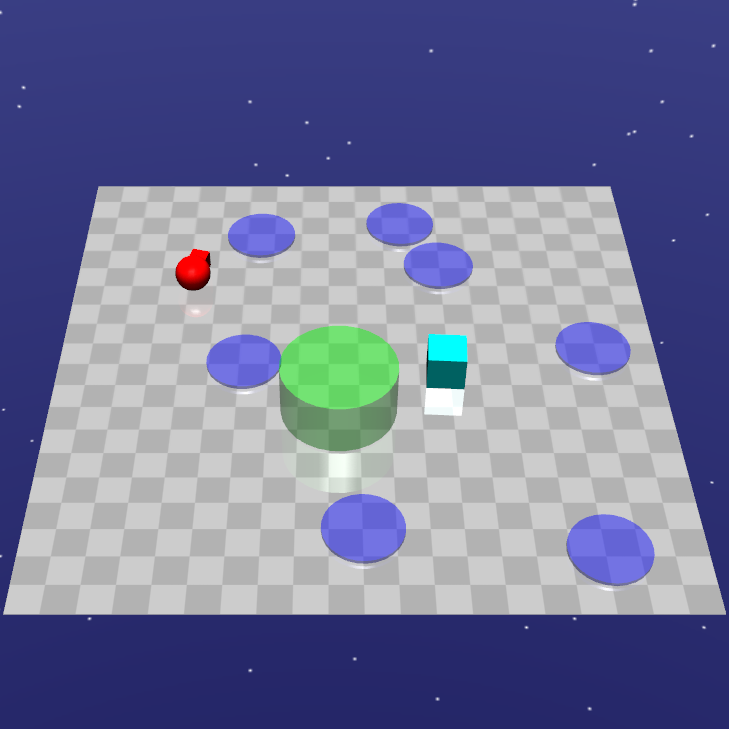}
  \caption{Point Goal1 Env}
  \label{fig:pg1}
\end{subfigure}%
\begin{subfigure}{.25\textwidth}
  \centering
  \includegraphics[height=2.8cm, trim={0 2cm 0 4cm},clip]{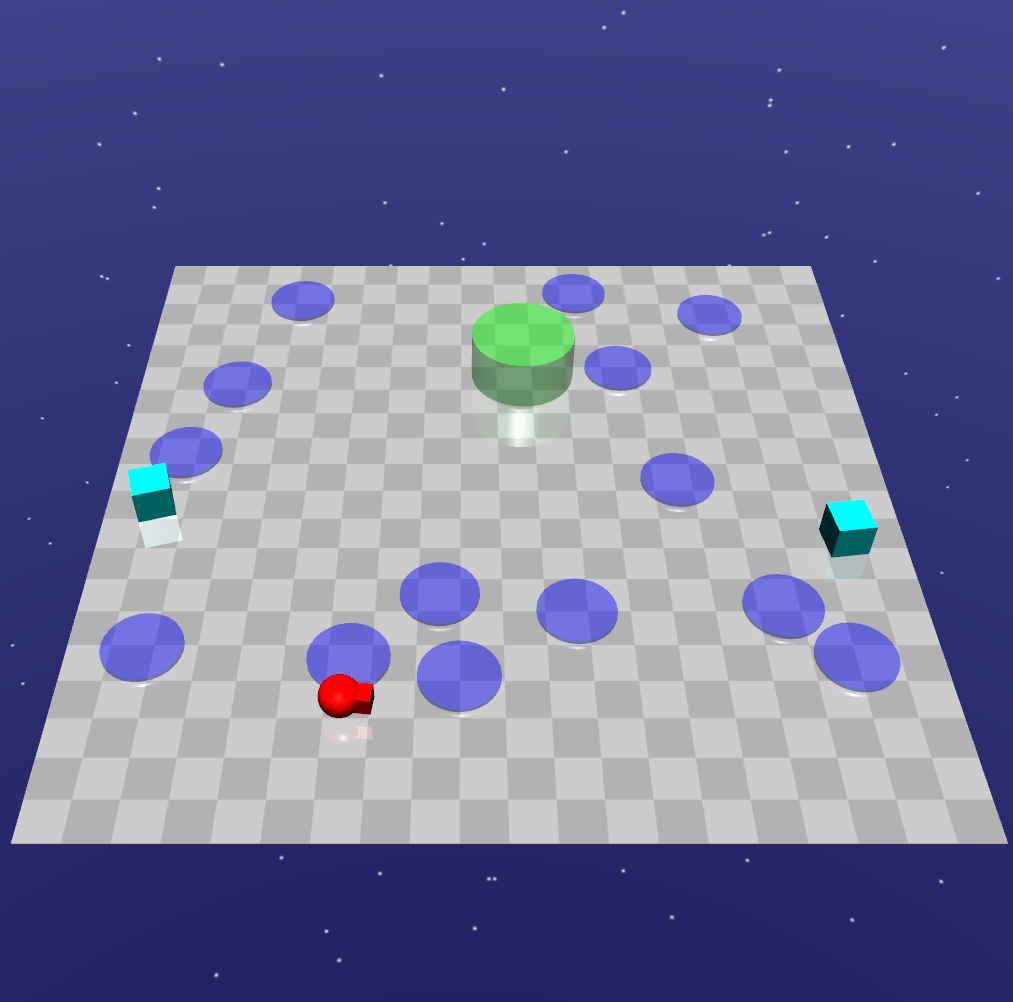}
  \caption{Point Goal2 Env}
  \label{fig:pg2}
\end{subfigure}%
\vskip\baselineskip
\begin{subfigure}{.25\textwidth}
  \centering
  \includegraphics[height=2.8cm, trim={0 1.5cm 0 3cm},clip]{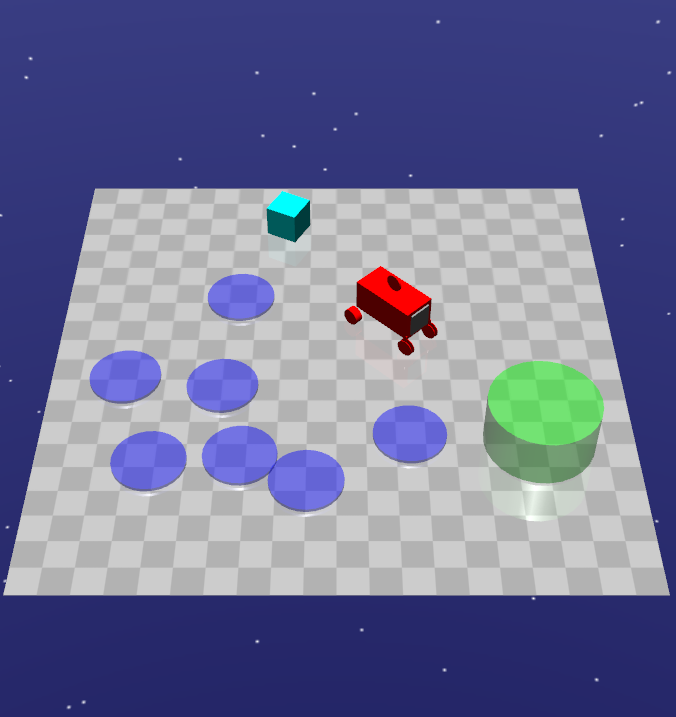}
  \caption{Car Goal1 Env}
  \label{fig:cg1}
\end{subfigure}%
\begin{subfigure}{.25\textwidth}
  \centering
  \includegraphics[height=2.8cm, trim={0 1cm 0 2.2cm},clip]{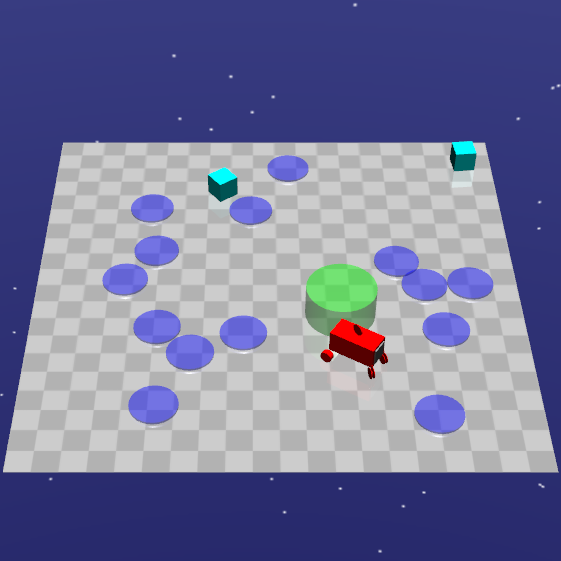}
  \caption{Car Goal2 Env}
  \label{fig:cg2}
\end{subfigure}
\setlength{\belowcaptionskip}{0pt}
\caption{Experiment Environments}
\label{fig:env}
%\vspace{-7mm}
% \end{figure*}
\end{center}

\subsection{Baselines and Experiment Setting}
\label{sec:baselines}

\begin{figure*}[t]
\centering
\includegraphics[width=0.94\textwidth]{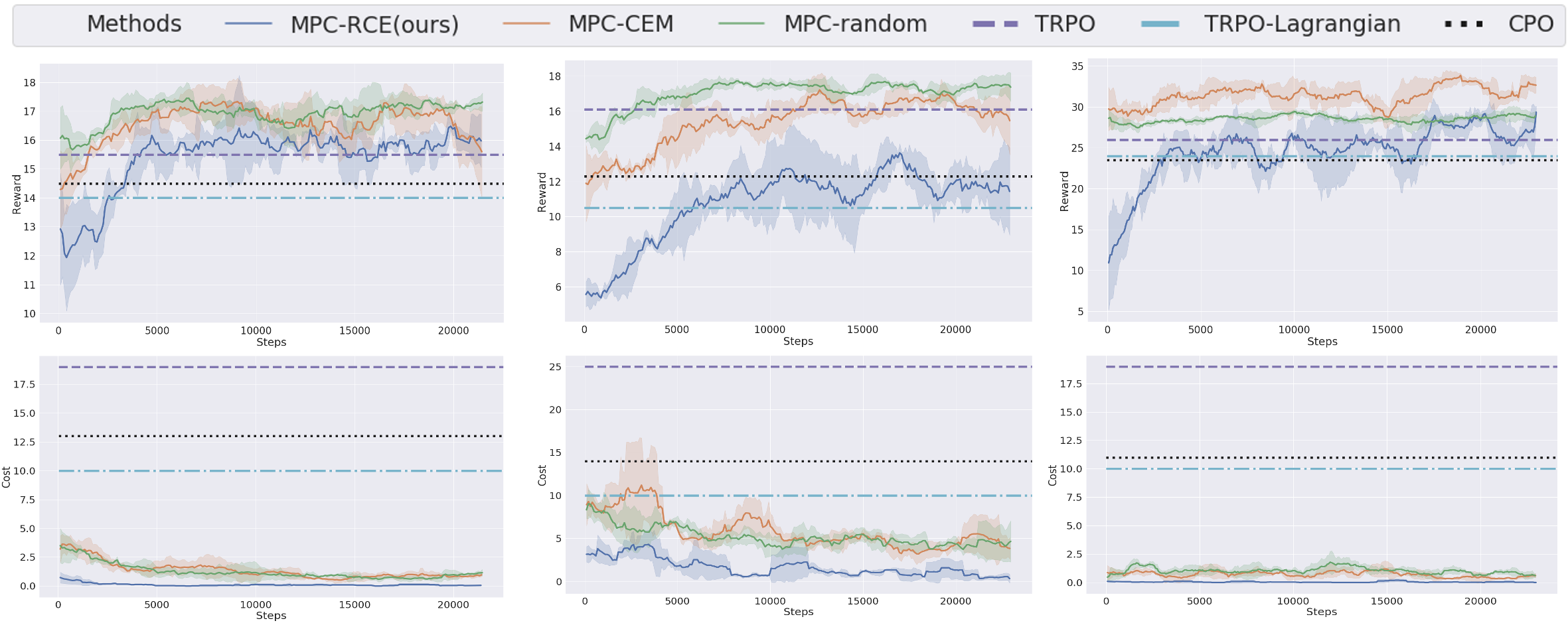}
\caption{Learning curves. The upper row figures are the reward trends and the lower row figures are the cost trends. From left to right the tasks are: Point Goal1, Point Goal2, and Car Goal1.}
\label{fig:training}
\vspace{-5mm}
\end{figure*}

\textbf{Baselines.} We use the official baseline methods provided by the Safety Gym environment~\cite{ray2019benchmarking}. The trust region policy optimization (TRPO) algorithm~\cite{schulman2015trust} is an unconstrained baseline that could give us intuition about the performance and constraint violations when we only care about the reward and not the cost of violating constraints. We use the Lagrangian version of TRPO (TRPO-Lagrangian) and the constrained policy optimization (CPO)~\cite{achiam2017constrained} as two model-free constrained reinforcement learning baselines, which we have introduced in section~\ref{sec:related}. 
%We observe that the proximal policy optimization (PPO) and its Lagrangian version perform similarly to TRPO and TRPO-Lagrangian, so we will not present them in our experiment result to make the figure more clear. 
For model-based constrained RL baselines, we use the MPC-random and MPC-CEM methods that are introduced in section~\ref{sec:mbbaseline}. %To distinguish from our uncertainty-aware solution, the two methods use the mean of the predictive distribution generated by our ensemble dynamic model and do not consider the prediction uncertainty. 

\textbf{Metrics and comparison.} We follow the metrics and comparison method in the Safety Gym paper~\cite{ray2019benchmarking}. We compare different approaches in terms of episodic accumulated reward and episodic cost, which is defined as the total constraint violation number in each episode. \textbf{Method $A$ is better than $B$ if $A$'s episodic cost is lower than $B$'s. If their costs are the same, then the one with higher episodic reward is better.} We also compare the sample efficiency based on the number of total interactions needed to converge, and compare the total cost during training.

\textbf{Training.} We use the same hyper-parameters for the model-based methods (MPC-RCE, MPC-CEM, and MPC-random) and the same hyper-parameters for the model-free baselines provided by the Safety Gym official benchmark~\cite{ray2019benchmarking}. For each algorithm, we evaluate them for each task with 3 different random seeds. For the detail about hyper-parameters used in our experiments, please refer to the appendix~\ref{sec:app} and code. The link of code is presented in the abstract.

Since RL agents learn by trial and error and we assume no prior knowledge of the environment, it is inevitable to violate the constraints in the early stage of training. However, our goal is to reduce the unsafe samples as much as possible because collecting unsafe data could be expensive in some cases. Such setting is practical in real-world applications. Here we provide two examples: 1. The cost to enter an ‘unsafe’ zone is not deadly but is not desired. For example, a tomato-picking-up robot damages tomatoes, which is not harmful to itself and people, but we still regard it as unsafe because it should be avoided in reality. 2. We can train the robot by historical unsafe data (such as public dataset) or in the simulator, which is not harmful during training. Then we can deploy trained constrained policies to real-world situations.

\subsection{Results}

% The training figures of reward and constraint violation number along with the interaction steps are shown in Fig.~\ref{fig:training}.
Fig.~\ref{fig:training} shows the learning curves of reward and constraint violation cost in Point Goal1, Point Goal2, and Car Goal1 tasks. The reward and cost are averaged among 3 experiments with the same hyper-parameters but different random seeds. The solid line is the mean value, and the light shade represents the area within one standard deviation. We use dashed lines to represent the value at convergence for model-free approaches, as they require several orders of magnitude more interaction steps to obtain satisfactory performance. Taking the Point Goal1 environment as an example, the model-based approaches converge after $5\mathrm{e}{3}$ steps with $5\mathrm{e}{3}$ initial data collected using the random policy, while TRPO-L requires $4\mathrm{e}{6}$ steps, which means our approach is 40 times more sample efficient compared to model-free methods.
 
From the figure, it is apparent that our MPC-RCE approach learns the underlying constraint function very quickly to avoid unsafe behaviors during the exploration and achieves the lowest constraint violation rate, though its reward is slightly lower than other methods. \textbf{It is reasonable because the best policy to maximize the task reward is to ignore the constraints and let the robot go straight towards the goal.} So MPC-CEM and MPC-random sacrifice the safety constraints satisfaction performance to obtain the gain on task reward. A more intuitive demonstration could be found in our supplementary video. However, as we introduced in section~\ref{sec:baselines}, we first compare the constraint satisfaction performance of different methods, and if they are the same, then we compare the task rewards. Therefore, our MPC-RCE is the best among all baselines because its constraint violation counts are always less than other methods, while the task rewards are comparable to other relatively 'unsafe' approaches.

For constrained model-free methods, a target cost value is required to be set in advance; it is set to be 10 in Fig.~\ref{fig:training}. We observed that if we set the target constraint violation rates to be a small value, the reward will decrease dramatically. In addition, these model-free methods constantly violate constraints during training to optimize the policy, which is not efficient and could be dangerous in practice. More experimental results and discussion on constrained model-free RL are presented in the appendix~\ref{sec:app}.

\begin{table}[]
\centering
\caption{Comparison of total constraint violation number for the first 10000 training steps.}
\begin{tabular}{|l|c|c|c|}
\hline
\backslashbox{Task}{Method}  & MPC-RCE       & MPC-CEM        & MPC-random     \\ \hline
Point Goal 1 & \textbf{16.00 }  & 184.33  & 169.0   \\ \hline
Point Goal 2 & \textbf{231.00 } & 746.00  & 600.67  \\ \hline
Car Goal 1   & \textbf{7.00 }   & 103.67  & 139.33  \\ \hline
Car Goal 2   & \textbf{96.00 }  & 578.00   & 509.33  \\ \hline
\end{tabular}
\label{tab:compare}
\vspace{-1mm}
\end{table}

Table~\ref{tab:compare} demonstrates the constraint satisfaction performance during the training procedure. We compare the total number of constraint violations for the first 10,000 steps of training. For constrained model-free RL methods, the total number of constraint violations is several orders of magnitude larger than model-based approaches, so we do not list the result here. From the table, we can see our approach achieves much lower cost than the other methods for the first 10,000 training steps, which means the MPC-RCE agent requires the minimum number of samples to converge and is able to explore the environment in a safer way. As far as we are aware, our method can achieve the best constraint satisfaction performance in these Safety Gym tasks.

\subsection{Why is RCE better than CEM?}
As we have shown in the results, RCE can achieve better constraint satisfaction performance than CEM, although they use the same dynamics model, constraint model, and hyper-parameters. Here we provide a short explanation to show that CEM is more likely to converge to unsafe action sequences than RCE when the agent is close to the dangerous areas. Denote the elite sample threshold as $k$. Consider a $t\text{-th}$ iteration during the optimization procedure. Suppose that in the sampled action sequences, only $q < \frac{k}{2}$ samples are safe, and the remaining samples will cause constraint violations, which is likely to happen when the agent is close to unsafe areas. For RCE method, the feasible set selection phase will help to discard all the unsafe samples in the elites, so that only the remaining safe samples will be used to update the sampling distribution at the $(t+1)$-th iteration. However, for the CEM method that simply adds cost penalties to the reward for unsafe samples, all the $k$ samples will be used to calculate the mean. Therefore, the mean of the $(t+1)\text{-th}$ sampling distribution will be biased towards unsafe samples, which causes more unsafe samples in the $(t+1)\text{-th}$ iteration.

%===============================================================================

\section{Conclusion}
\label{sec:conclusion}
We introduce a simple yet effective constrained model-based RL algorithm without any prior assumptions on the system dynamics or the constraint function. We propose the robust cross-entropy method (RCE) to optimize the action under the MPC framework in light of the model uncertainty and underlying constraints. Our method is evaluated in the Safety Gym environment and achieves better constraint satisfaction while maintaining good task performance compared with other constrained RL baselines. 
%However, performing online optimization at every time-step could be computationally expensive in high-dimensional environments. 
One potential future direction is to improve the cost model to provide a measure of risk towards longer horizons rather than only a single step prediction.

% \addtolength{\textheight}{-10cm}   % This command serves to balance the column lengths

%%%%%%%%%%%%%%%%%%%%%%%%%%%%%%%%%%%%%%%%%%%%%%%%%%%%%%%%%%%%%%%%%%%%%%%%%%%%%%%%

%%%%%%%%%%%%%%%%%%%%%%%%%%%%%%%%%%%%%%%%%%%%%%%%%%%%%%%%%%%%%%%%%%%%%%%%%%%%%%%%

%%%%%%%%%%%%%%%%%%%%%%%%%%%%%%%%%%%%%%%%%%%%%%%%%%%%%%%%%%%%%%%%%%%%%%%%%%%%%%%%


\section*{ACKNOWLEDGMENT}
The authors acknowledge the support from the Manufacturing Futures Initiative at Carnegie Mellon University made possible by the Richard King Mellon Foundation which supported Zuxin Liu and Baiming Chen. Zuxin Liu is also in part supported by Carnegie Mellon University's College of Engineering Moonshot Award for Autonomous Technologies for Livability and Sustainability (ATLAS) and by Carnegie Mellon University's Mobility21 National University Transportation Center, which is sponsored by the US Department of Transportation. The authors would like to thank Keegan Harris, Peide Huang and Zixin Ye for their valuable suggestions of this paper. 

\bibliographystyle{IEEEtran}
\bibliography{main}
%%%%%%%%%%%%%%%%%%%%%%%%%%%%%%%%%%%%%%%%%%%%%%%%%%%%%%%%%%%%%%%%%%%%%%%%%%%%%%%%

\section*{APPENDIX}
\label{sec:app}
\subsection{Discussion about the Safety Gym Environment and the Task Setting}

Safety Gym environments use the MuJoCo physics engine as the backbone simulator. Each environment and task is inspired by a practical safety issue in robotics control.
The observation spaces used in the original Safety Gym environment includes standard robot sensors (accelerometer, gyroscope, magnetometer, and velocimeter) and pseudo-lidar (each lidar sensor perceives objects of a single kind and is computed by filling bins with appropriate values). The observation space used in our approach is different from the default Safety Gym options in that we pre-process the sensor data to get rid of some noisy and unstable sensors, such as the $z$-axis data of accelerometer. We use the relative coordinates of the perceived objects instead of the pseudo-lidar readings because the former representation is more friendly to dynamics model learning, which is important for model-based RL.

%Both robots used in our experiment have two-dimensional continuous action spaces and all actions are linearly scaled to $[-1, +1]$. We also performed careful hand-tuning of some MuJoCo actuator parameters during sensor analysis, since robust and responsive control is critical to robot operations in both the simulation environment and the real world. 

Our work in the Safety Gym environment has implications for real-world applications. The Goal task in our experiment resembles the setting of the delivery robot and other domestic robots, where the robot has to navigate around static obstacles such as furniture to reach the goal. Additionally, since the state representation in our experiments is directly derived from sensor information and the control input of our environment to the robot is very similar to that of real-world situations, our model-based RL approach in the simulation environment could serve as an important pre-training for the real-world applications. Given that a certain amount of unsafe data is required to train our model, it would be unrealistic to have the real robot repeatedly violate the constraints to collect such data. Therefore, the training in the simulator is an important step for the model to be transferable to real wold safety-critical applications.

\begin{figure*}[h]
\centering
\begin{subfigure}{.48\textwidth}
  \centering
  \includegraphics[width=1\linewidth]{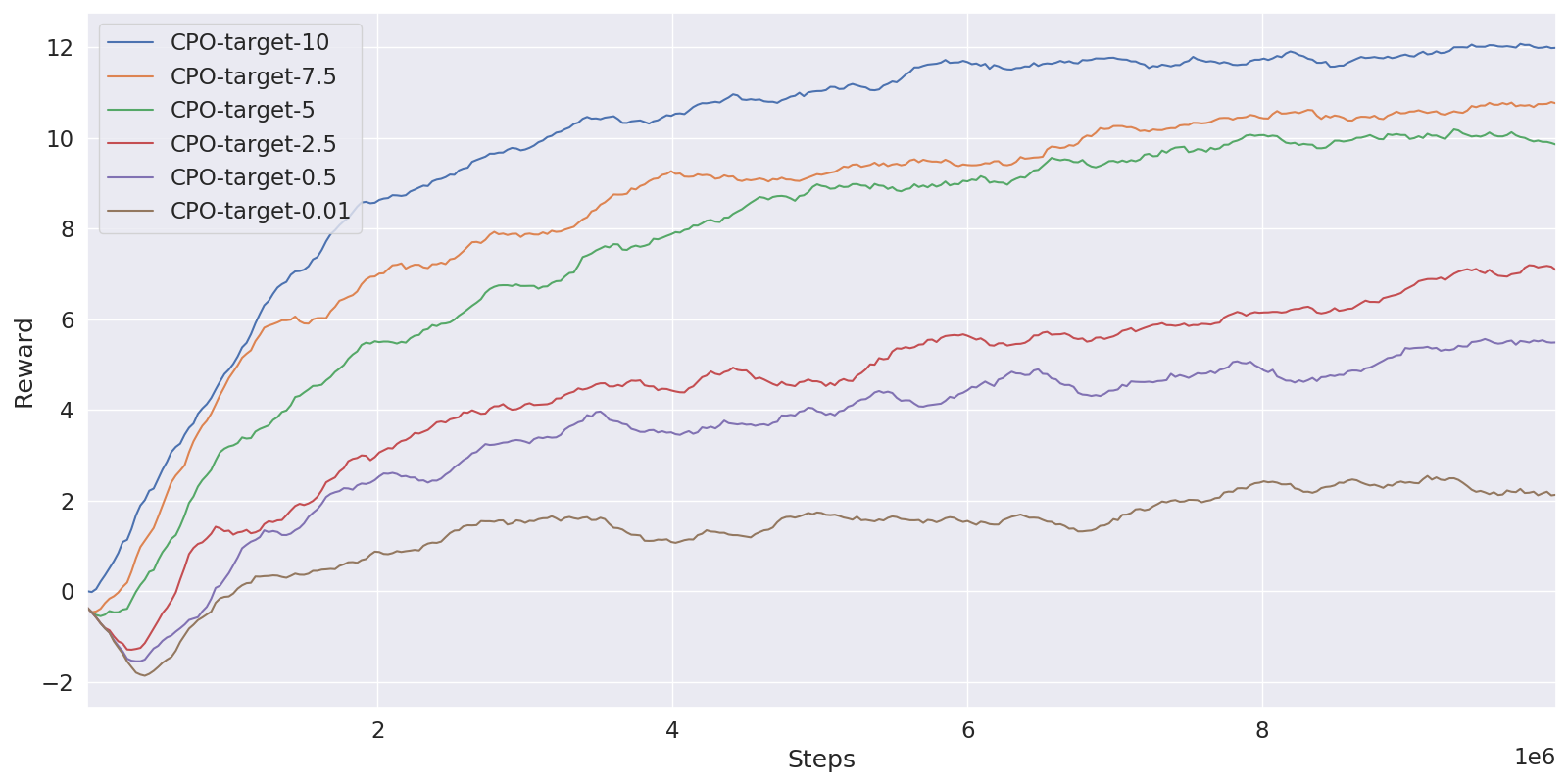}
\end{subfigure}%
\begin{subfigure}{.48\textwidth}
  \centering
  \includegraphics[width=1\linewidth]{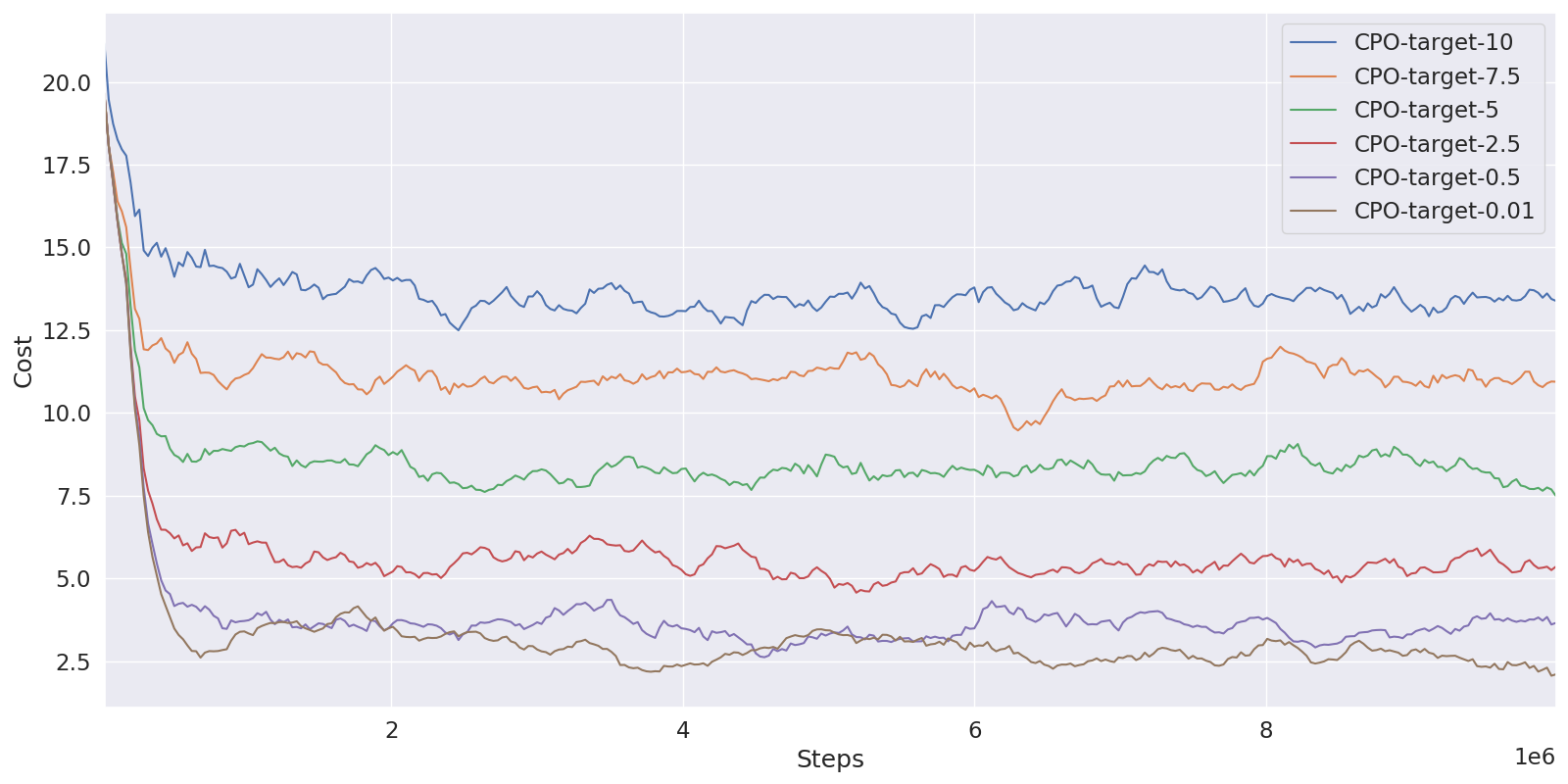}
\end{subfigure}%
\caption{Learning curves of reward and cost for CPO with different target cost value.}
 \label{fig:cpo}
\end{figure*}

\begin{figure*}[h]
\centering
\begin{subfigure}{.48\textwidth}
  \centering
  \includegraphics[width=1\linewidth]{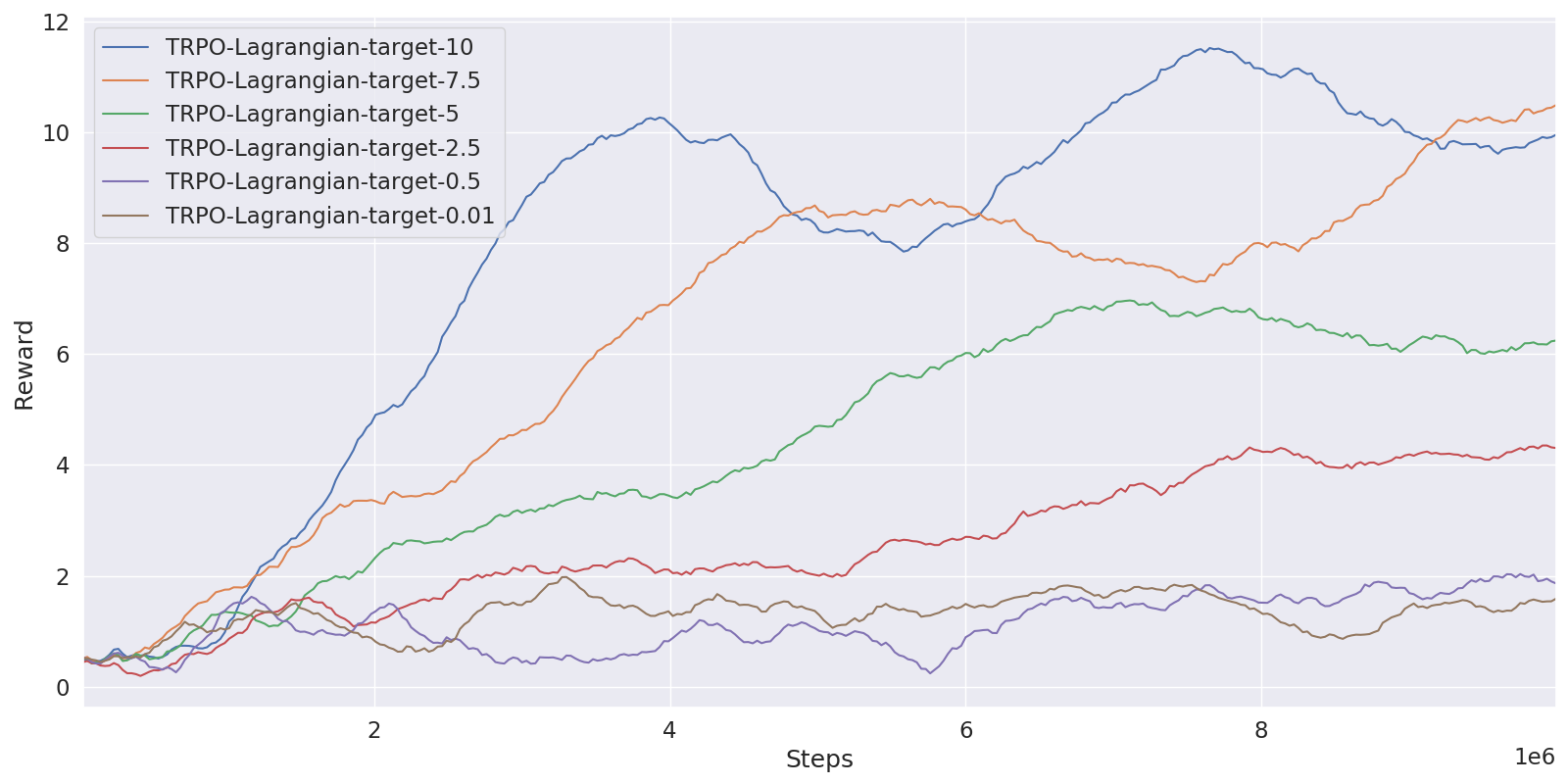}
\end{subfigure}%
\begin{subfigure}{.48\textwidth}
  \centering
  \includegraphics[width=1\linewidth]{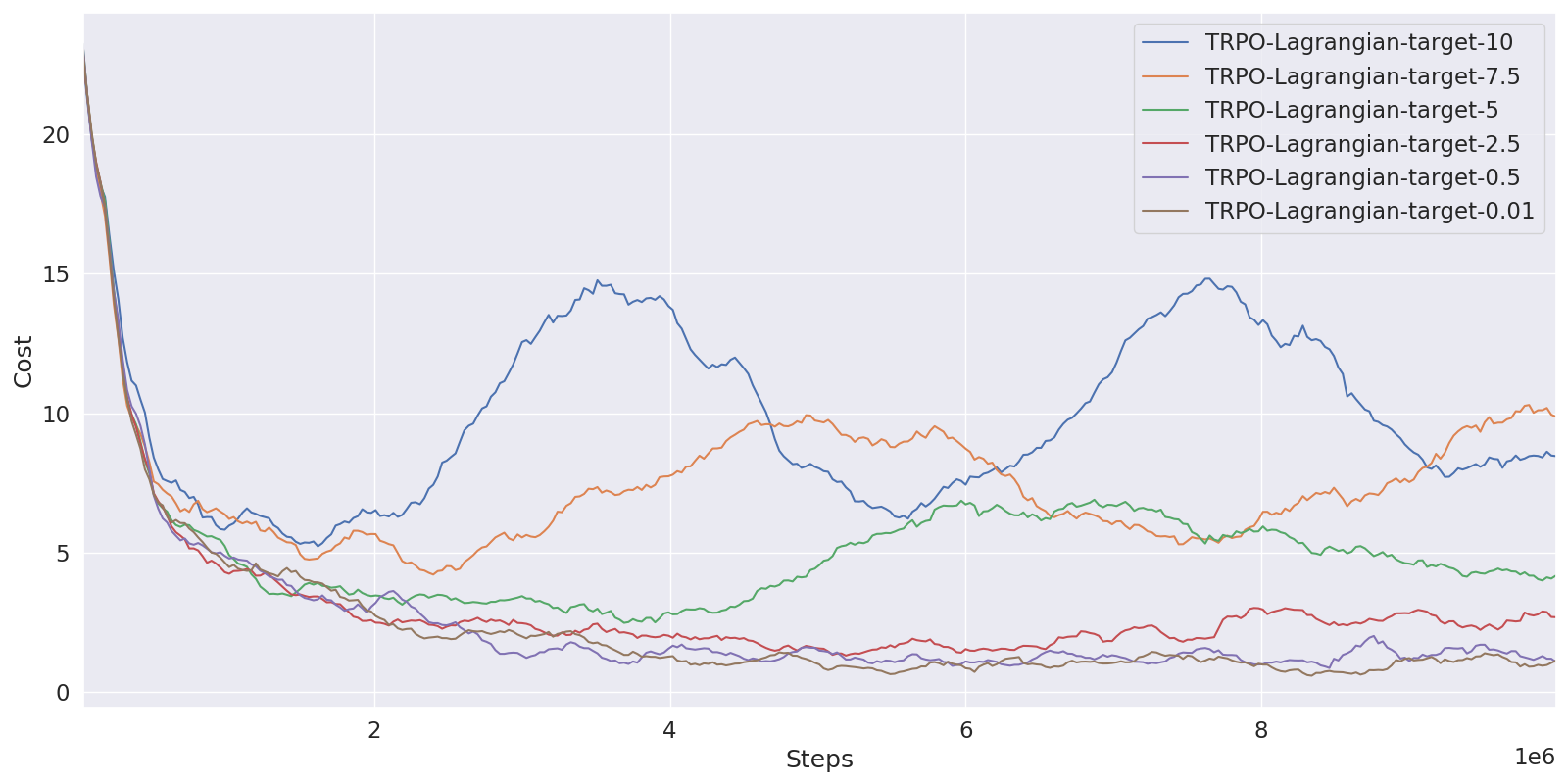}
\end{subfigure}%
\caption{Learning curves of reward and cost for TRPO-Lagrangian with different target cost value.}
 \label{fig:trpo}
\end{figure*}

\subsection{Training Detail}
\label{sec:param}

\textbf{Dynamics model}:
We use the same architecture and hyper-parameters of each neural network in the ensemble dynamics model. Each neural network is of 3 layers with ReLU activation and each layer is of 1024 neurons. All the training parameters for one task are the same for MPC-RCE, MPC-CEM, and MPC-random. The batch size is 256, the learning rate is 0.001, the training epochs are 70, and the optimizer is Adam. The ensemble number is 4 for Point robot-related tasks and is 5 for Car robot-related tasks. Each neural network model in the ensemble is trained with $80\%$ of the training data to prevent overfitting.

\textbf{LightGBM classifier}: We use LightGBM to predict the constraint violation given a state in our RCE method and all the model-based baselines. We use the default \texttt{gdbt} boosting type and 400 base estimators. Each base estimator has a maximum depth of 8 and 12 leaves. The learning rate is 0.3 and all other hyper-parameters are the default value.

\textbf{RCE, CEM, and random optimizer}: We use the same hyper-parameters for RCE and CEM except that RCE has a discount value of $\gamma=0.98$ for reward and discount value of $\beta=0.4$ for cost while CEM only has one discount value $\gamma=0.98$ for the combination of reward and cost. We sample $N=500$ solutions for each iteration of RCE and CEM and select top $k=12$ elite samples to estimate the distribution parameters for the next iteration. If the iteration number exceeds 8 or the sum of the variance of elite samples is less than $\epsilon=0.01$, the optimization procedure stops and returns the best solution that has been found so far. To fairly compare with RCE and CEM, we use 5000 samples for the random shooting method so that the maximum number of samples is at the same order of magnitude. The planning horizon is $T=8$ for all methods.

\textbf{TRPO, TRPO-Lagrangian, and CPO}: We use the same hyper-parameters offered in the open-sourced code from the baseline method for the Safety Gym simulation environment~\cite{ray2019benchmarking}. All hyperparameters are kept the same for all three model-free baseline methods. The actor-critic neural network model has 2 linear layers of 256 hidden neurons in each. The discount factor $\gamma=0.99$. The target cost limit is 10 with penalty term $\lambda$ initialized to be $ 1$ and a penalty term learning rate of $0.05$. The target KL divergence is $0.01$, and for the value function learning, the learning rate is $0.001$ with 80 iterations. For each experiment, the total number of environment interactions is $1\mathrm{e}{7}$ and $3\mathrm{e}{4}$ steps for each training epoch. More hyper-parameters information can be found in the source code.

\subsection{More Results}
\label{sec:figures}

\textbf{The influence of the target cost value for TRPO-Lagrangian and CPO}. Since the target cost limit value must be set in advance before training, we empirically study the performance of TRPO-Lagrangian and CPO with different target values in the Point Goal2 environment. The learning curves are shown in Fig~\ref{fig:cpo} and Fig~\ref{fig:trpo}. We can see that the task performance is negatively correlated with the target cost, and there is a dramatic task performance drop if we limit the target cost to a small value. Compared with model-based approaches, CPO and TRPO-Lagrangian can hardly achieve comparable task performance with the same level of constraint violation rate.

\end{document}

%% file: math_command.tex
\usepackage{amsmath,amsfonts,bm}

\newcommand{\Xcal}{\mathcal{X}}

\newcommand{\Tcal}{\mathcal{T}}
\newcommand{\Ncal}{\mathcal{N}}
\newcommand{\Scal}{\mathcal{S}}
\newcommand{\Dcal}{\mathcal{D}}

\newcommand{\Acal}{\mathcal{A}}

\def\eqref#1{equation~\ref{#1}}
\def\1{\bm{1}}

\DeclareMathAlphabet{\mathsfit}{\encodingdefault}{\sfdefault}{m}{sl}
\SetMathAlphabet{\mathsfit}{bold}{\encodingdefault}{\sfdefault}{bx}{n}